# Longitudinal Characterization of Personalized Hydration Dynamics Using a Facial-Video Digital Hydration Biomarker

Janindu Athukorala, D.M.J. Prathapa, Thanveer Ahamed, Yudara Kularathne
Hii.Health, Singapore

## Abstract

**Background.** Hydration is usually assessed with single measurements and managed with population fluid-intake targets, although body water changes continuously and responses to fluid may differ between people. A facial-video digital hydration biomarker has been shown to detect hydration-state transitions; whether the same tool can follow hydration within an individual over hours has not been examined.

**Objective.** To determine whether the biomarker tracks within-person hydration changes across the day in a physiologically plausible way, to quantify how hydration dynamics and responses to the same fluid input differ between individuals, and to explore whether individual response patterns could inform personalized hydration protocols.

**Methods.** Five adult volunteers completed two experiments. In a 12-hour longitudinal study, each participant completed a water day and an electrolyte day with an identical 2.0 L fluid schedule and a 30-second facial video every 30 minutes (250 videos). In a 2-hour calibration study, a baseline video after waking was followed by 500 mL of an isotonic electrolyte drink and videos every 30 minutes (25 videos). Outcomes were score trajectories, within-person responses to drinking, each participant's scan-to-scan variation and least significant change, response magnitude and time course, time in range (score $\geq$ 50) and stability. Exploratory analyses examined whether responses depended on the score before drinking and compared the two study days from their first scan.

**Key findings.** The first scan of the day, taken after the overnight period without fluid, was the lowest score of that day on 7 of 10 days. Across all 50 study-fluid intakes, the mean post-drink change (+4.0 points) exceeded that expected when timing was removed by permutation ($p < 0.001$), and the response depended on the state in which fluid was taken: none of the 21 intakes taken when the score was below 50 was followed by a fall, and their mean rise was about twice that expected from regression toward the mean (+11.3 vs +5.7 points; $p < 0.001$), whereas intakes taken at or above 50 produced no rise beyond chance. After the same 500 mL drink, every participant's score rose (mean change over 2 hours +5.4 to +12.3 points), but the peak came at 30, 60 or 90 minutes depending on the participant and ranged from +7.2 to +22.6 points. Under an identical 2.0 L schedule, time in range varied from 16% to 100%, and from similar morning scores the score rose further on the electrolyte day than on the water day in all five participants (mean difference +7.7 points).

**Conclusion.** The facial-video hydration biomarker can characterize hydration longitudinally within individuals. Hydration trajectories, variability and responses to the same fluid input differ markedly between people, suggesting that personal calibration may be necessary before individualized hydration protocols can be defined. Larger prospective studies are required.

## 1. Introduction

Hydration is usually treated as a static, population-level concept. Fluid-intake guidance is expressed as population reference values, for example total water intakes of 2.0 L/day for women and 2.5 L/day for men in Europe [1] and adequate intakes of 2.7 and 3.7 L/day in the United States [2]. Hydration status itself is most often judged from a single measurement, such as plasma osmolality, a urine index or a change in body mass, each of which describes one moment in time and carries practical constraints [3, 4]. Even well-established markers show substantial biological variation, which limits the value of a fixed cut-off when interpreting an individual measurement [5].

New early evidence shows that a facial-video digital hydration biomarker can detect a change in hydration state [6]. The biomarker is derived from remote photoplethysmography (rPPG)-associated signals, which capture cardiovascular-related colour changes of facial skin with an ordinary camera [7, 8], and it requires only a short smartphone video. This makes repeated, non-invasive measurement practical in everyday settings.

Detecting a large before-and-after change is a necessary first step, but daily hydration involves smaller, repeated changes. The next question is whether the same tool can track hydration longitudinally within an individual: how the score moves across a day, how it responds to each drink, and how stable it remains between drinks.

There are good reasons to expect hydration to be both dynamic and highly personal. Body water is turned over continuously: it is lost through urine, breath and skin and replaced through drinking and food, with renal water handling regulated by arginine vasopressin and thirst [4, 9]. Ingested fluid must leave the stomach and be absorbed before it changes body-water distribution, and the speed of these steps depends on drink volume and composition [10]. Beverages also differ in how much of the ingested fluid is retained over the following hours [11]. Between people, habitual fluid intake is associated with different urinary hydration profiles [12], and sweat rate and sweat sodium concentration vary several-fold [13].

These differences point to a need for personalized hydration protocols. Sports medicine already recommends individualized fluid plans based on measured sweat losses [14], and personalized strategies have improved fluid balance compared with drinking at will [15]. Outside the laboratory, however, there is no practical way to observe how a given person responds to fluid. In personalized nutrition, measuring individual responses showed that the same meal can produce very different glycaemic responses between people and allowed advice to move beyond population averages [16]; repeated measurements within single individuals are central to that approach [17]. In this study we apply the same logic to hydration, using two complementary experiments in the same five individuals to address three objectives.

### Study objectives

- **Objective 1 – longitudinal measurement.** Show that the digital hydration biomarker can track hydration changes longitudinally within a person, with patterns that are physiologically plausible.
- **Objective 2 – hydration is personal.** Show that hydration responses differ substantially between individuals, even after similar fluid inputs, using the 12-hour and 2-hour data.
- **Objective 3 – toward personal calibration.** Explore whether personal response patterns can support future individualized hydration protocols. This objective is preliminary; larger prospective studies are required before optimal dosing or ranges can be defined.

**Secondary concept.** Longitudinal measures such as trajectory, stability and time in range may be more useful than an isolated hydration measurement or a daily fluid-volume target.

## 2. Methods

### 2.1 Study design

The study comprised two complementary experiments carried out by the same five participants (Fig. 1).

### *A. 12-hour longitudinal study*

Each participant completed two 12-hour monitoring days: a water day followed by an electrolyte day. On both days, a facial video was recorded before any food or fluid intake (first scan, T0) and then every 30 minutes until T+12 h, giving 25 videos per day. Total fluid intake was standardized at 2.0 L: 1.5 L of study fluid at five fixed times and 0.5 L of non-study fluid (tea or coffee) with three meals at fixed times (Table 1). On the water day the study fluid was plain water; on the electrolyte day it was a commercially available, ready-to-drink isotonic electrolyte drink supplied in 500 mL bottles (sodium 12.2 mEq/L, potassium 8.4 mEq/L, calcium 7.3 mEq/L, magnesium 4.9 mEq/L; Table 2). The drink also contains 5.8 g of carbohydrate per 100 mL, all as sugar, and the sweetener sucralose; according to the manufacturer, its osmolality matches that of blood. Each participant followed the identical schedule, timed from their first scan, on both days. Participants were instructed to record each video with the same smartphone, seated and at rest, with the face fully visible and without speaking. Meals, non-study drinks, symptoms and medication use on each day were recorded.

**Table 1.** Standardized fluid and meal schedule of the 12-hour study (identical on both days).

| Time from first scan | Study fluid | Non-study fluid (tea/coffee) | Meal |
|---|---|---|---|
| T0 (after first scan) | 400 mL | – | – |
| T+30 min | – | 150 mL | Breakfast |
| T+2 h 30 min | 300 mL | – | – |
| T+4 h | – | 150 mL | Lunch |
| T+5 h 30 min | 300 mL | – | – |
| T+8 h 30 min | 250 mL | – | – |
| T+10 h | – | 200 mL | Dinner |
| T+11 h | 250 mL | – | – |
| **Total** | **1,500 mL** | **500 mL** | |

Study fluid was plain water on the water day and the isotonic electrolyte drink on the electrolyte day. A facial video was recorded every 30 min from T0 to T+12 h (25 per day).

**Table 2.** Electrolyte composition of the isotonic electrolyte drink.

| Electrolyte | Concentration (mEq/L) | Amount per litre (mg) |
|---|---|---|
| Sodium ($Na^+$) | 12.2 | 280 |
| Potassium ($K^+$) | 8.4 | 330 |
| Calcium ($Ca^{2+}$) | 7.3 | 147 |
| Magnesium ($Mg^{2+}$) | 4.9 | 59 |

Values are from the product label. Salts: sodium chloride; potassium citrate, phosphate and sorbate; calcium chloride; magnesium sulphate.

### *B. 2-hour calibration study*

On a separate morning, each participant recorded a baseline facial video after waking, then drank 500 mL of the same isotonic electrolyte drink and recorded further videos at +30, +60, +90 and +120 minutes (five videos per participant). No other food or drink was consumed during the 2-hour window.

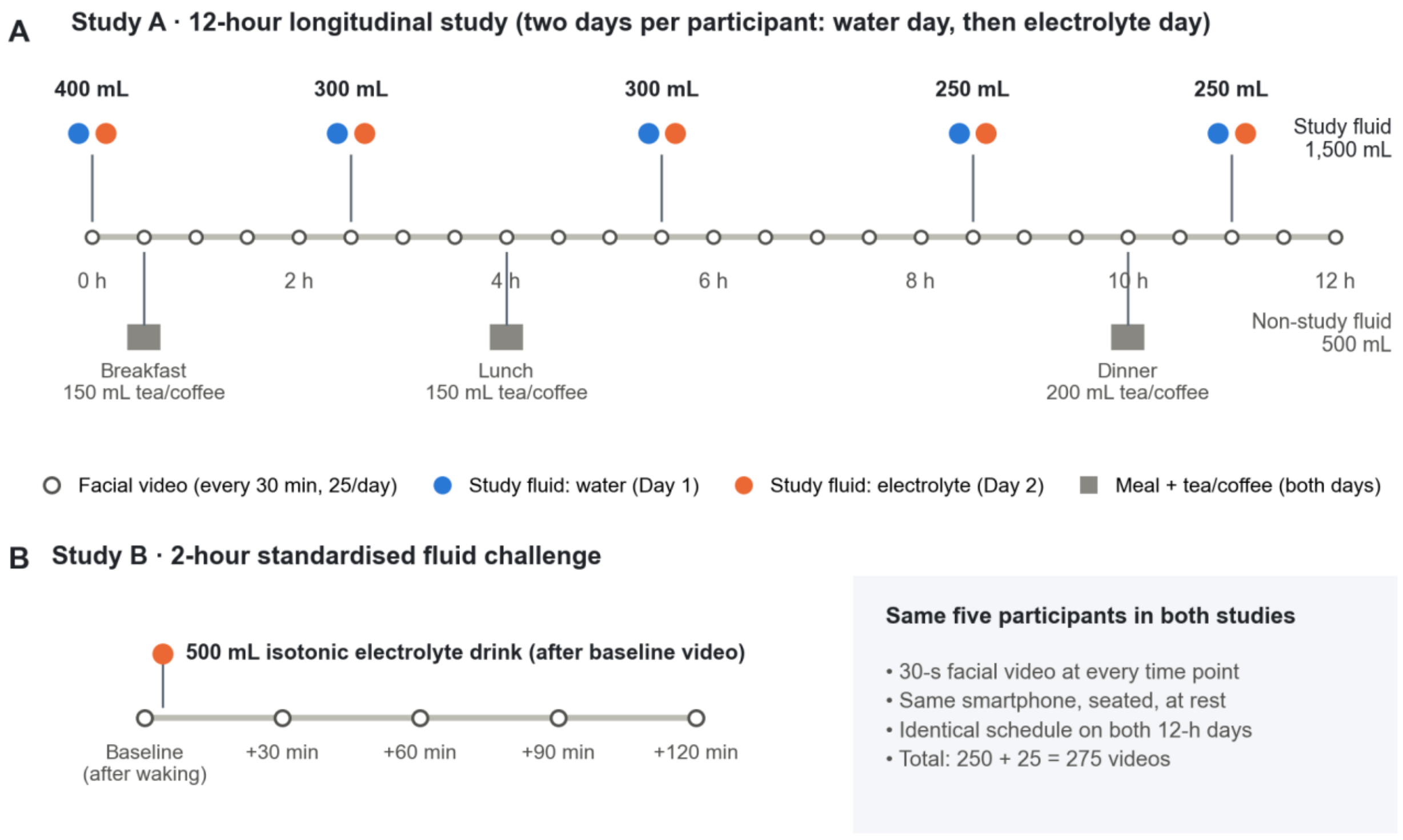

**Fig. 1.** Study design. (A) 12-hour longitudinal study: each participant completed a water day and an electrolyte day with an identical schedule of study fluid (circles; 1,500 mL), meals with tea or coffee (squares; 500 mL) and facial videos every 30 minutes. (B) 2-hour calibration study: baseline video after waking, 500 mL of the isotonic electrolyte drink, then videos every 30 minutes.

## 2.2 Participants

Five adult volunteers (three men and two women, aged 26–30 years) took part in both experiments. Participants were not selected for a particular health status: body size ranged from underweight to obese (body mass index 18.2–32.6 kg/m$^2$), and one participant had chronic medical conditions treated with regular medication and was involved under the supervision of physician (Table 3). Height, body weight, health background, regular medication, menstrual cycle and symptoms on each study day were recorded by questionnaire. All participants gave informed consent to take part and to the publication of anonymized health information. Participants are identified as Participant 1 to Participant 5.

## 2.3 Digital hydration biomarker

The biomarker produces a hydration score from a 30-second facial video recorded with a standard smartphone camera. Facial skin regions are tracked through the video, and rPPG-associated signals are extracted and analysed by a proprietary deep-learning model of the type developed for video-based physiological measurement [18]. The score ranges from 0 to 100; higher values indicate a state more consistent with recent fluid intake, and 50 is the operational boundary between states more consistent with a hydrated or a under hydrated condition. All 275 videos were scored with a single, fixed version of the score, and no data from the present studies were used to train or tune the model. The score is intended as a lifestyle digital biomarker for tracking the direction and pattern of hydration change; it is not a diagnostic test for dehydration.

## 2.4 Longitudinal outcomes

The following outcomes were defined before analysis.

- **Score trajectory:** the sequence of 25 scores on each 12-hour day and five scores in the 2-hour study.
- **Within-person change:** for each of the five study-fluid intakes per 12-hour day (50 intakes in total), the drink response was the mean of the scores 30 and 60 minutes after intake minus the score immediately before intake. In the 2-hour study, change was expressed relative to the baseline video.
- **Between-person variability:** each participant's scan-to-scan variation, $\sigma$, estimated robustly from the 48 consecutive-scan differences ($\Delta$) across both 12-hour days as $\sigma = 1.4826 \times \text{median}|\Delta| / \sqrt{2}$; the standard deviation (SD) of scores within each day; and a decomposition of total score variance into between-participant, between-day and within-day components.
- **Meaningful change:** following the reliable-change and least-significant-change principle [19, 20], a change was considered meaningful only if it exceeded, with 95% confidence, the change expected from scan-to-scan variation alone: $\text{LSC} = 1.96 \times \sqrt{2} \times \sigma$ when two single scans were compared, and $1.96 \times \sqrt{1.5} \times \sigma$ when a single pre-drink scan was compared with the mean of two post-drink scans. Because consecutive-scan differences also contain real physiological change, $\sigma$, and therefore the LSC, is a conservative (upper-bound) estimate.
- **Response magnitude and time course (2-hour study):** peak change from baseline, time to peak, change at 120 minutes, and mean change over 0–120 minutes (area under the change curve divided by 120 minutes). Responses were classified as a sustained rise (above the LSC at two or more time points including 120 minutes), a transient rise (above the LSC at one or more time points but not at 120 minutes) or, when no time point exceeded the LSC, a rise within the LSC (positive mean change over 0–120 minutes) or no rise; meaningful falls were recorded separately.
- **Time in range and stability:** time in range was the percentage of a day's scans with a score of 50 or above; also recorded were the longest continuous period at or above 50 and the number of exits below 50. Stability was the mean absolute change between consecutive scans. To compare a single measurement with the whole day, we also calculated the probability that one scan taken at a random time disagreed with the day's predominant state (the smaller of time in range and its complement).

Additional analyses, defined after inspection of the data, examined whether drink responses depended on the pre-drink score (below 50 vs 50 or above) and compared the two study days within each participant from their first scan, using the change from the first scan so that differences in starting level did not affect the comparison. These analyses are reported as exploratory.

### 2.5 Statistical approach

The analysis was primarily descriptive and exploratory, emphasized individual trajectories and within-person comparisons, and avoided population-level claims from a small sample. Three tests were used to guard against over-interpretation. First, a one-sided sign test asked whether the score rose after the first drink of the day more often than the 50% expected by chance. Second, a permutation test [21] asked whether the scores after study-fluid intake were higher than expected from the day's own distribution of scores: the order of scans within each day was shuffled 20,000 times, keeping each day's values but removing their timing relative to drinks, and the observed mean drink response across all 50 intakes was compared with the shuffled distribution. Because each shuffled day retains the same scores, this test also accounts for regression toward the mean after unusually low pre-drink scans. Third, the Fligner–Killeen test [22] compared the spread of consecutive-scan differences between participants. In the exploratory analyses, the one-sided Wilcoxon signed-rank test, which also uses the size of each change, supplemented the sign test and compared the first scan with the mean of the rest of the day. The permutation test was repeated with responses defined over 30–90 minutes, and separately for intakes taken below 50 and at or above 50; for each group, the observed mean response was compared with the mean response of the intakes that fell in the same group in the shuffled data, which contain the same regression toward the mean. Within-participant differences between the two study days, and the mean change over the 2-hour challenge, were summarized with t-based 95% confidence intervals and tested with an exact sign-flip test across the five participants (smallest attainable one-sided

p = 0.031). P-values are reported without adjustment and are interpreted as exploratory. Analyses were performed in Python 3.12 (NumPy, SciPy and pandas); data and analysis code are archived with the study files.

## 3. Results

### 3.1 Participant characteristics

All 275 scheduled videos (250 in the 12-hour study and 25 in the 2-hour study) were available and included in the analysis. Participant characteristics are summarized in Table 3. The group was deliberately heterogeneous: one participant had a normal body mass index and no health conditions, two were overweight, one was obese and one was underweight. Participant 3 had several chronic conditions treated with regular medication, and Participant 5 had migraine and experienced a migraine episode on the water day. Three participants reported discomfort on the electrolyte day from drinking only the electrolyte drink, without plain water.

**Table 3.** Participant characteristics.

| Participant | Sex | Age (y) | Height (cm) | Weight (kg) | BMI ($kg/m^2$) | BMI category (WHO) | Health context |
|---|---|---|---|---|---|---|---|
| Participant 1 | M | 26 | 170.2 | 62 | 21.4 | Normal | No medical conditions; no regular medication. |
| Participant 2 | M | 29 | 176.8 | 102 | 32.6 | Obese (class I) | No medical conditions; no regular medication. |
| Participant 3 | M | 30 | 173.7 | 82 | 27.2 | Overweight | Chronic kidney condition, chronic obstructive pulmonary disease, hypertension, sleep difficulties and anxiety; regular antihypertensive, antiplatelet and anxiolytic medication. |
| Participant 4 | F | 30 | 157.5 | 63 | 25.4 | Overweight | No medical conditions; no regular medication. Regular 28-day menstrual cycle; no menstrual symptoms on study days. |
| Participant 5 | F | 29 | 169.0 | 52 | 18.2 | Underweight | Migraine on regular preventive medication. Water day: migraine episode, analgesic taken. Electrolyte day: menstruation, mild headache. |

Health context was self-reported.

### 3.2 12-hour longitudinal hydration patterns

The hydration score changed throughout the day on all ten monitoring days (Fig. 2), with within-day ranges from 19.8 to 36.3 points; on every day the range exceeded the participant's own least significant change. Four features of the trajectories were consistent with expected hydration physiology (Fig. 3).

First, the morning scan was usually the low point of the day. The first scan of the day, taken after the overnight period without fluid, was the lowest of that day's 25 scores on 7 of 10 days (0.4 days expected by chance) and fell in the lowest quarter on 9 of 10 days. It was below 50 on 8 of 10 days and averaged 41.5 points, against 54.5 points for the rest of the day (higher later in the day on 9 of 10 days; Wilcoxon signed-rank p = 0.002).

Second, the first drink of the day was followed by a rise whenever the day began below 50 (Fig. 3A). On all eight such days the score rose (median change +14.5 points, range +1.8 to +24.0; one-sided sign test p = 0.004) and reached 50 or above within one hour. The two days that began at or above 50 were both Participant 1's, and on these days the score did not rise (−1.9 and −0.6 points). Across all ten days the score rose on eight (median change +13.8 points; one-sided sign test p = 0.055; Wilcoxon signed-rank test p = 0.007). In four cases the first-drink response exceeded the participant's own least significant change (Participant 2 on both days, +14.1 and +22.0 points; Participant 3, electrolyte day, +24.0 points; Participant 5, electrolyte day, +16.5 points).

Third, drinking was followed by a higher score more often than expected by chance. Across all 50 study-fluid intakes the mean drink response was +4.0 points, compared with 0.0 ± 1.1 points (mean ± SD) when the timing of scans was removed by permutation ($p < 0.001$; Fig. 3B); the result was unchanged when responses were defined over 30–90 minutes (+3.5 points; $p < 0.001$). Six intakes produced a rise beyond the personal least significant change, against 1.6 expected by chance ($p = 0.002$). Responses to the later, smaller drinks (250–300 mL) were smaller: the score rose after 24 of 40 later intakes, with a mean change of +2.1 points, and 2 of these 40 rises exceeded the least significant change.

Fourth, the response depended on the state in which fluid was taken (Fig. 3C). None of the 21 intakes taken when the score was below 50 was followed by a fall: their mean response was +11.3 points, about twice the +5.7 points expected from regression toward the mean in the shuffled data ($p < 0.001$), and the pattern held both for the first drink of the day (8 intakes; +14.8 vs +5.5 points; $p = 0.022$) and for later drinks (13 intakes; +9.2 vs +5.7 points; $p = 0.041$). Intakes taken when the score was already at or above 50 were followed by no rise beyond chance (29 intakes; −1.3 vs −2.3 points expected; $p = 0.17$), and the only two falls beyond the least significant change followed intakes taken at high scores (61.0 and 61.2 points). Because most later drinks (27 of 40) were taken at or above 50, their average response was small.

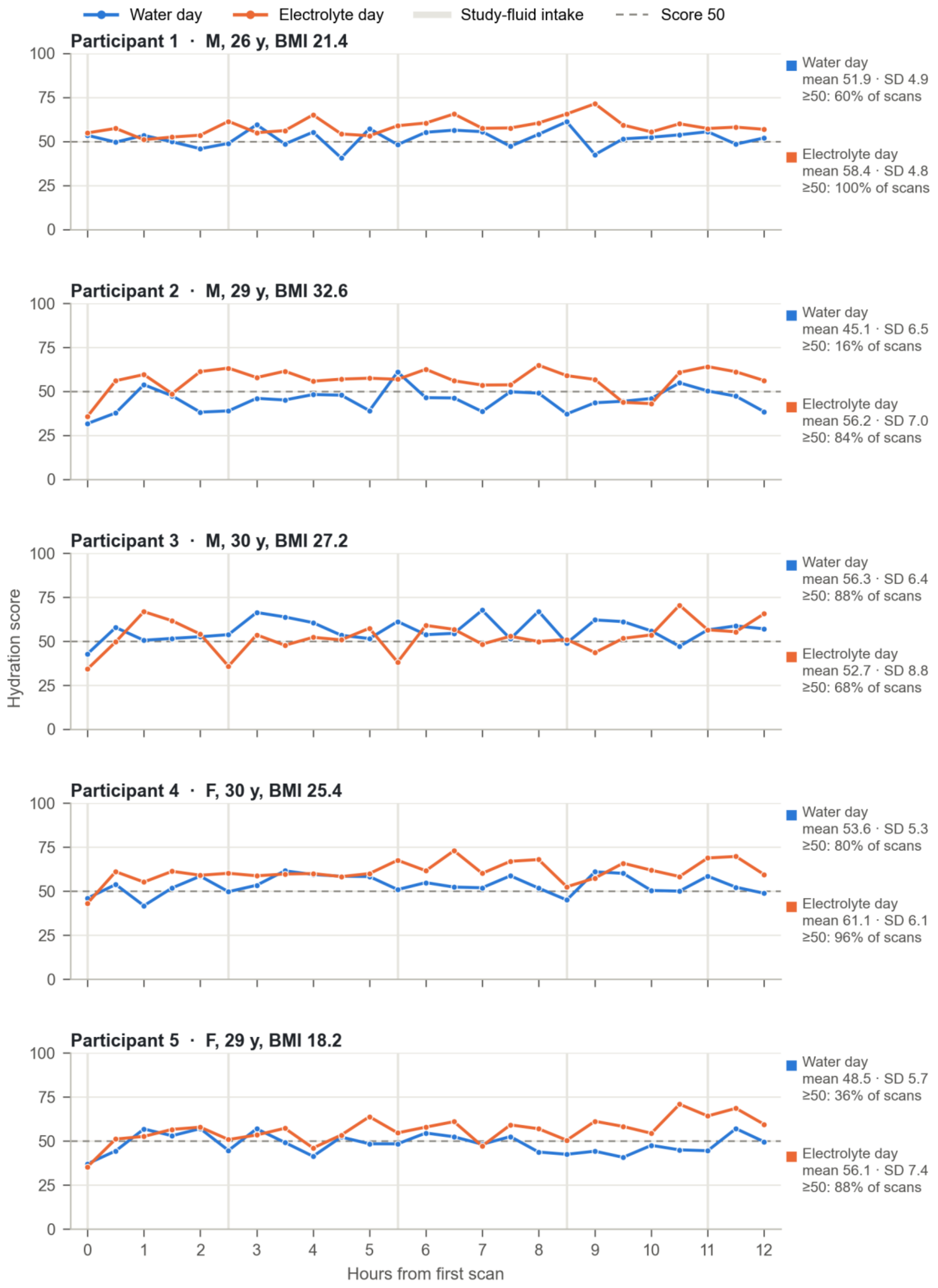

**Fig. 2.** 12-hour hydration-score trajectories for each participant on the water day (blue) and the electrolyte day (orange). Light vertical bands mark study-fluid intake (400, 300, 300, 250 and 250 mL); the dashed line marks a score of 50. Right: day mean, SD and percentage of scans at or above 50.

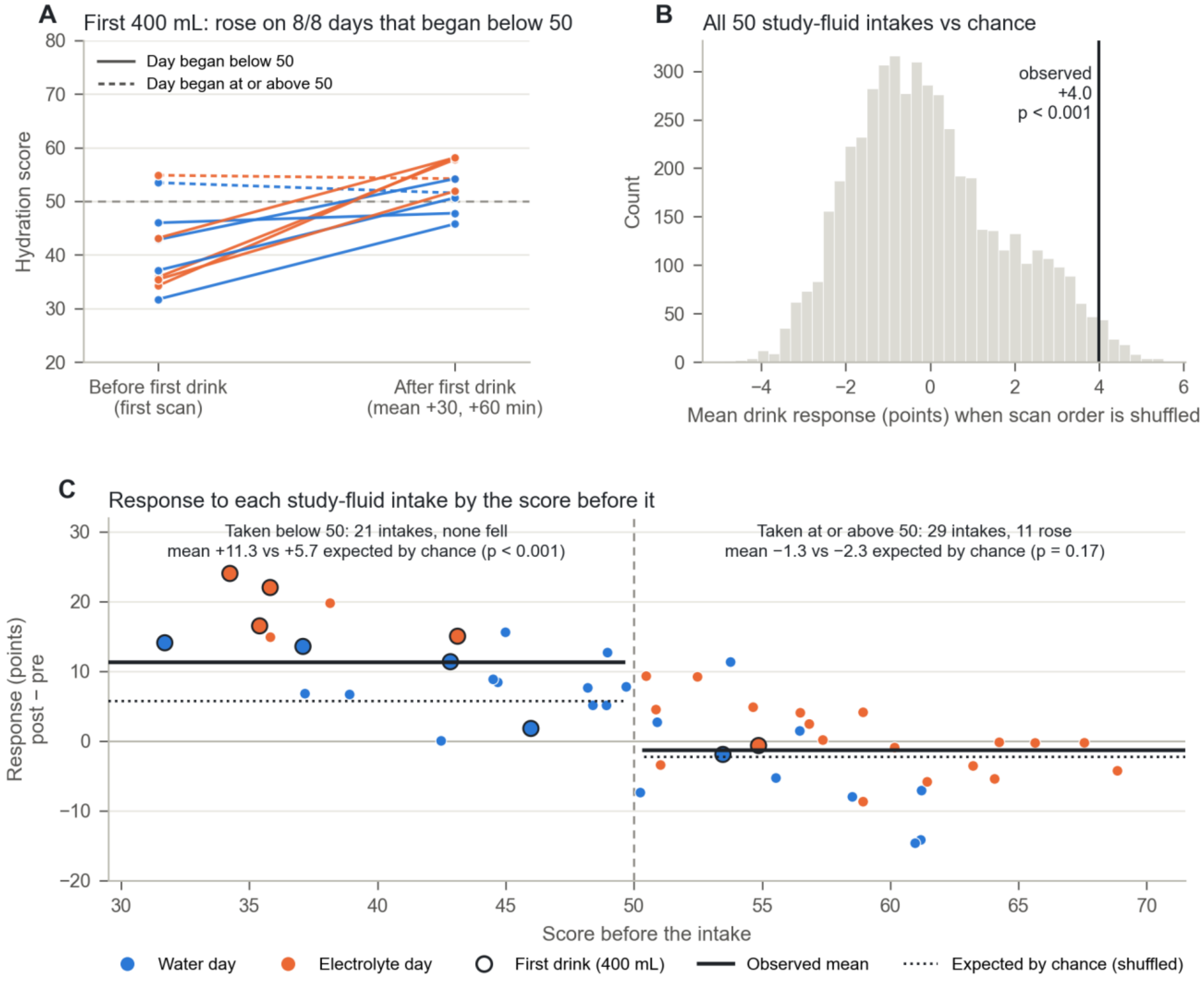

**Fig. 3.** Responses to drinking in the 12-hour study. (A) Score before the first drink of the day and the mean of the +30 and +60 min scores after it, for all ten days; dashed lines mark the two days that began at or above 50. (B) Mean drink response across all 50 study-fluid intakes compared with the distribution obtained when scan order is shuffled within each day (5,000 shuffles shown; the reported p-value uses 20,000). (C) Response to each intake plotted against the score before it; solid lines show the observed mean response of intakes taken below 50 and at or above 50, dotted lines the mean expected by chance in the shuffled data, and outlined markers the first drink of the day.

## 3.3 Hydration responses differ between individuals

The amount by which the score moved from one scan to the next differed between participants (Fig. 4A). Scan-to-scan variation ranged from 5.1 points (Participant 1) to 7.7 points (Participant 3) (Fligner–Killeen $\chi^2 = 13.5$, df = 4, $p = 0.009$). The corresponding least significant change between two scans ranged from 14.2 to 21.4 points, so a 15-point change exceeded measurement variation for Participant 1 but not for Participant 3. In contrast, participants differed little in their average level: 66.1% of the total variance in scores lay within days (over time), 24.9% between the two days of the same participant (that is, between the water and electrolyte days; see below) and only 9.0% between participants. What distinguished participants was therefore not their average score but how their score moved (Fig. 4B).

Curve shapes reflected these differences (Fig. 2). Participant 1 was the only participant to begin both days at or above 50 (53.5 and 54.9 points) and stayed within a band of about 20 points on both days; on the electrolyte day the score never fell below 50. Participant 2 began both days with the lowest scores of the group (31.7 and 35.8 points); on the water day the score remained below 50 for most of the day (time in range 16%), with pre-drink scores of 31.7, 38.9 and 37.2 points before three of the five drinks. Participant 3 varied most: on the electrolyte day the score fell to 34–38 points before each of the first three drinks and rose by 15 to 24 points after them. Participant 4 stayed mostly at or above 50 on both days (80% and 96% of scans). Participant 5 spent most of the water day, on which a migraine episode occurred, below 50 (time in range 36%), with every pre-drink score below 50, whereas on the electrolyte day the score was at or above 50 for 88% of the day.

The two study days also differed within participants (Fig. 7C, D). Before any study fluid, morning scores were similar on the two days (electrolyte minus water, mean −1.5 points; range −8.6 to +4.1). From that starting point, the score rose further on the electrolyte day than on the water day in all five participants (mean difference in change from the first scan +7.7 points; 95% CI 4.5 to 10.8; exact sign-flip p = 0.031), beginning with the first 400 mL drink (mean response +15.4 vs +7.8 points; larger on the electrolyte day in all five). Later drinks produced similar responses on both days (+2.0 and +2.2 points); the difference lay in how well the score was maintained between drinks. Before the four later drinks, the score had fallen below 50 on 11 of 20 occasions on the water days but on 2 of 20 on the electrolyte days, and time in range was higher on the electrolyte day in four of five participants (mean 87% vs 56%; Table 4, Fig. 7A). The exception was Participant 3: the rise from the first scan was larger on the electrolyte day, but that day began 8.6 points lower, and its mean score and time in range were not higher (52.7 vs 56.3 points; 68% vs 88%). Because the water day always preceded the electrolyte day, these differences should be confirmed in a randomized-order study.

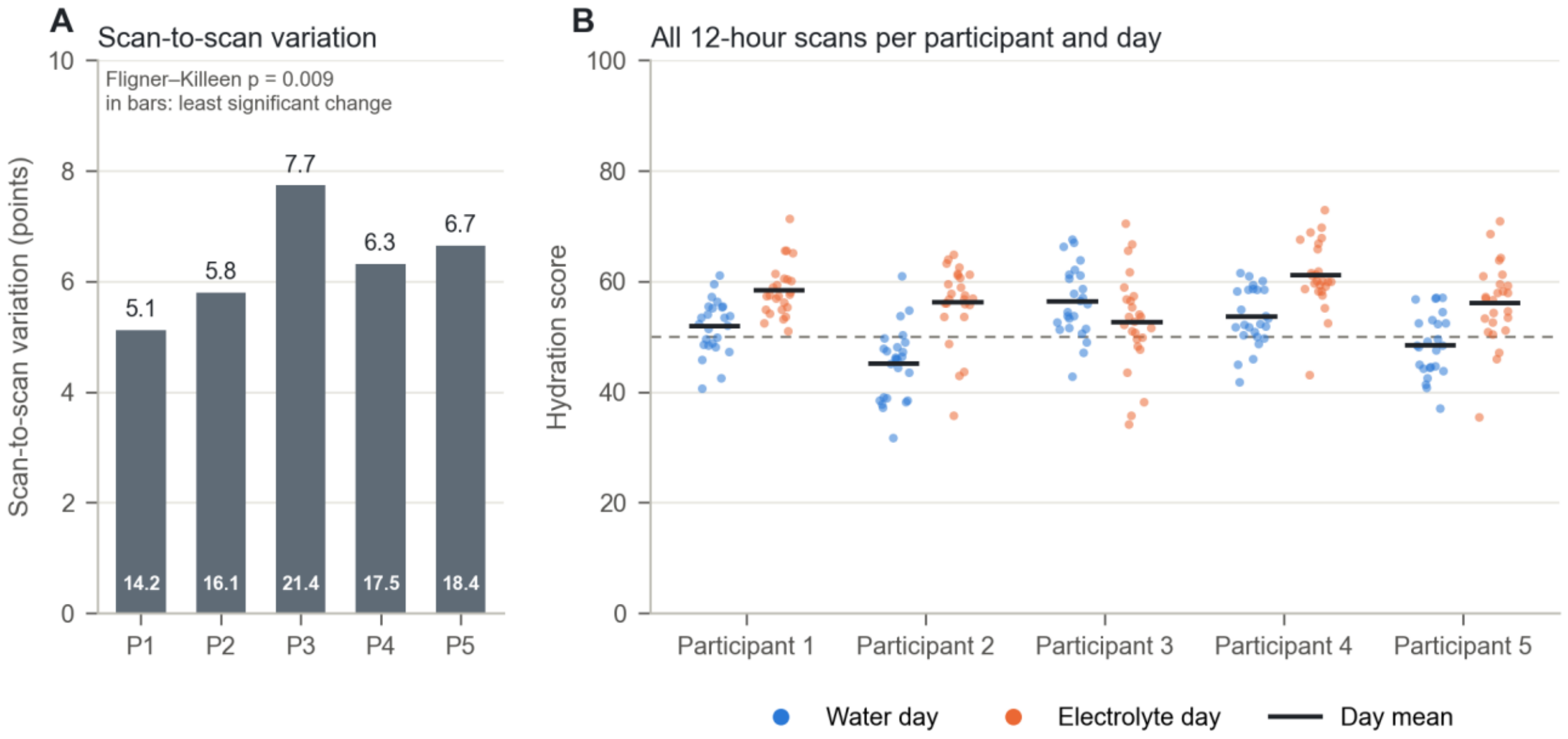

**Fig. 4.** Between-person differences. (A) Each participant's scan-to-scan variation, estimated from the 48 consecutive-scan differences across both 12-hour days; numbers in the bars give the corresponding least significant change between two scans. (B) All 12-hour scans for each participant and day; black bars show day means and the dashed line marks a score of 50.

**Table 4.** 12-hour longitudinal metrics for each participant and day.

| Participant | Day | Mean | SD | Range | Time in range (%) | Longest run ≥ 50 (h) | Exits < 50 | Stability | First scan | First-drink response |
|---|---|---|---|---|---|---|---|---|---|---|
| Participant 1 | Water | 51.9 | 4.9 | 40.7–61.2 | 60 | 1.5 | 8 | 6.7 | 53.5 | −1.9 |
| Participant 1 | Electrolyte | 58.4 | 4.7 | 51.0–71.4 | 100 | 12.0 | 0 | 4.5 | 54.9 | −0.6 |
| Participant 2 | Water | 45.1 | 6.5 | 31.7–61.0 | 16 | 0.5 | 3 | 6.7 | 31.7 | +14.1 |
| Participant 2 | Electrolyte | 56.2 | 7.0 | 35.8–64.8 | 84 | 7.0 | 2 | 6.0 | 35.8 | +22.0 |
| Participant 3 | Water | 56.3 | 6.4 | 42.8–67.6 | 88 | 7.5 | 2 | 7.3 | 42.8 | +11.4 |
| Participant 3 | Electrolyte | 52.7 | 8.8 | 34.2–70.5 | 68 | 2.5 | 6 | 9.1 | 34.2 | +24.0 |
| Participant 4 | Water | 53.6 | 5.3 | 41.7–61.6 | 80 | 5.0 | 4 | 5.9 | 46.0 | +1.8 |
| Participant 4 | Electrolyte | 61.1 | 6.1 | 43.1–73.0 | 96 | 11.5 | 0 | 6.0 | 43.1 | +15.0 |
| Participant 5 | Water | 48.5 | 5.7 | 37.1–57.1 | 36 | 1.0 | 6 | 6.0 | 37.1 | +13.5 |
| Participant 5 | Electrolyte | 56.1 | 7.4 | 35.4–70.9 | 88 | 4.5 | 2 | 7.1 | 35.4 | +16.5 |

Each day comprised 25 scans at 30-minute intervals from the first scan (T0) to T+12 h. Time in range: % of the 25 scans with a score ≥ 50. Longest run: the longest continuous period during which every scan was ≥ 50, measured from the first to the last of those scans (maximum 12 h). Stability: mean absolute change between consecutive scans (points). First scan: score before any food or fluid (T0). First-drink response: mean of +30 and +60 min scores minus the score before the first 400 mL drink.

### 3.4 2-hour calibration responses

Every participant's score rose after the same standardized fluid challenge, but the timing and size of the rise differed (Fig. 5, Table 5). Baseline scores after waking ranged from 35.9 to 55.3. Of the 20 post-drink scans, 19 were above the participant's baseline, and the mean change over the 2 hours was positive in all five participants. The peak came at 30 minutes in Participant 5 (+10.6 points), at 60 minutes in Participant 1 (+18.5 points) and at 90 minutes in Participants 2, 3 and 4 (+14.8, +22.6 and +7.2 points). Participant 1 rose earliest, exceeding the personal least significant change at 30 and 60 minutes. Participant 3 dipped at 30 minutes (−7.9 points), the only post-drink scan below baseline, and then rose beyond the least significant change at 90 minutes. Participant 2 reached 50 only at 90 minutes and returned to 40.1 by 120 minutes. In Participants 4 and 5 the score reached 50 by 30 minutes and then stayed close to that level, a rise within the personal least significant change.

Peak changes therefore ranged from +7.2 to +22.6 points, about a three-fold difference, with peaks at 30 to 90 minutes. Averaged across participants, the change from baseline was +6.2 points at 30 minutes, +10.5 at 60, +12.6 at 90 and +7.7 at 120 minutes, and it was positive in all five participants from 60 minutes onward (Fig. 5, last panel). Because the participants' least significant changes differed, a similar change in points carried different weight for different people: a rise of 15 points was beyond measurement variation for Participant 1 (least significant change 14.2 points) but within it for the other four participants (16.1 to 21.4 points). Averaged over the full 2 hours (area under the change curve divided by 120 minutes), the mean change from baseline ranged from +5.4 points (Participant 4) to +12.3 points (Participant 1), with a group mean of +8.3 points (95% CI 5.2 to 11.4; Table 5).

The comparable event in the 12-hour study, the first drink on the electrolyte day (400 mL of the same drink, also given after the overnight period), did not reproduce the individual pattern of the 2-hour challenge (Table 5). Two differences between the settings may explain this. First, in the 12-hour study breakfast with tea or coffee followed 30 minutes after the first drink, whereas the 2-hour challenge was taken without food. Second, the starting state differed: Participant 1 began both 12-hour days above 50 and showed no first-drink rise, but began the 2-hour challenge just below 50 (48.7 points) and showed the earliest rise beyond the least significant change. Starting state was not the only factor, because Participant 3 began the challenge above 50 (55.3 points) and still rose beyond the least significant change after an initial dip. With one challenge per person, these observations are preliminary.

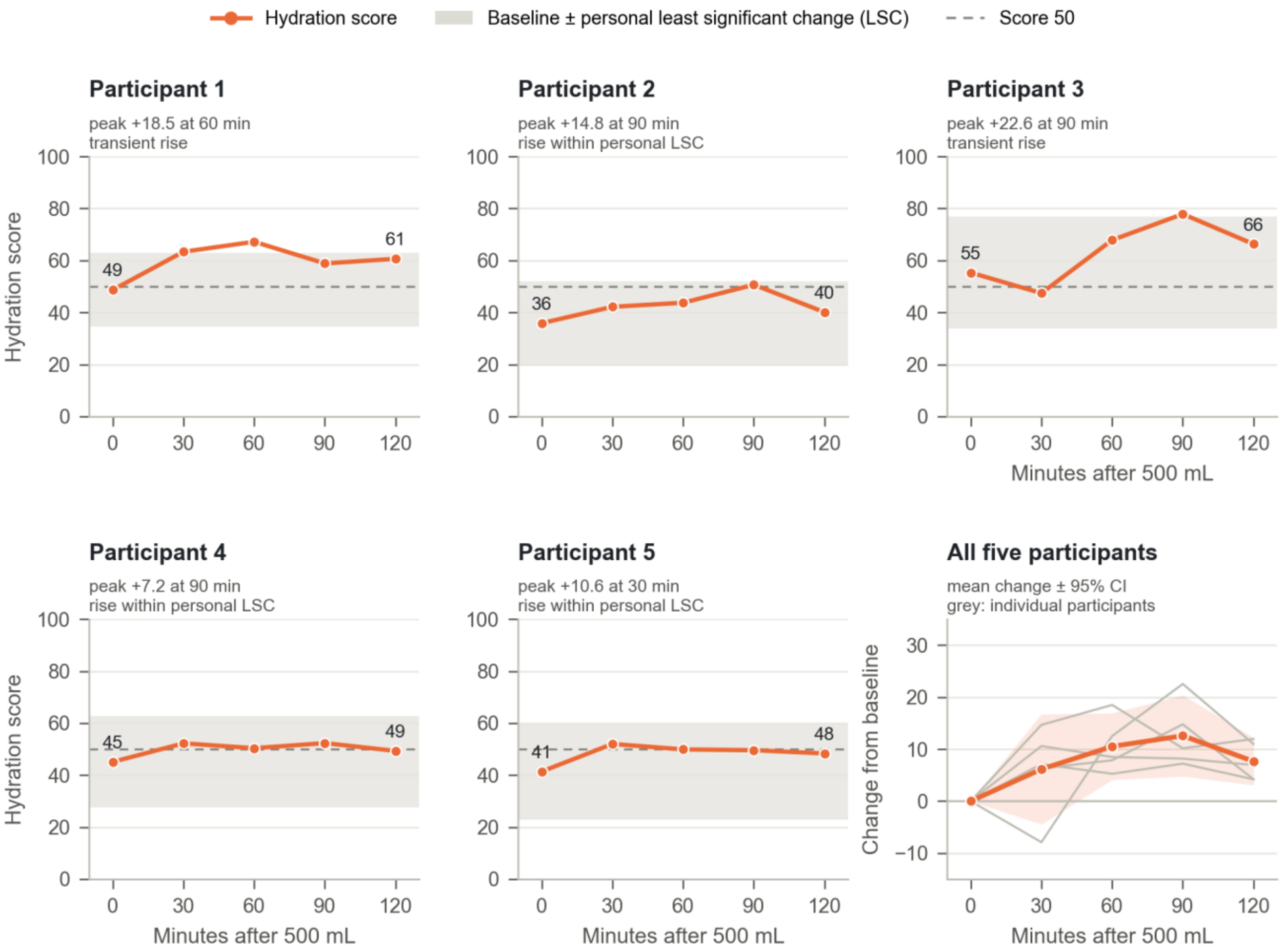

**Fig. 5.** Responses to the standardized 2-hour challenge (500 mL isotonic electrolyte drink after a baseline video on waking). Grey band: baseline ± the participant's least significant change between two scans, derived from their 12-hour data. Dashed line: score 50. Last panel: mean change from baseline across the five participants (± 95% CI); grey lines, individual participants.

**Table 5.** Responses in the 2-hour calibration study and the comparable event in the 12-hour study.

| Participant | Baseline | Δ 30 | Δ 60 | Δ 90 | Δ 120 | Mean Δ 0–120 | Peak Δ (min) | LSC | Classification vs LSC | 12-h first drink, electrolyte day |
|---|---|---|---|---|---|---|---|---|---|---|
| Participant 1 | 48.7 | +14.7 | +18.5 | +10.2 | +12.0 | +12.3 | +18.5 (60) | 14.2 | Transient rise | −0.6 |
| Participant 2 | 35.9 | +6.3 | +7.9 | +14.8 | +4.2 | +7.8 | +14.8 (90) | 16.1 | Rise within LSC | +22.0 |
| Participant 3 | 55.3 | −7.9 | +12.5 | +22.6 | +11.0 | +8.2 | +22.6 (90) | 21.4 | Transient rise | +24.0 |
| Participant 4 | 45.1 | +7.1 | +5.3 | +7.2 | +4.2 | +5.4 | +7.2 (90) | 17.5 | Rise within LSC | +15.0 |
| Participant 5 | 41.4 | +10.6 | +8.5 | +8.2 | +7.0 | +7.7 | +10.6 (30) | 18.4 | Rise within LSC | +16.5 |

Δ, change from baseline (points) at each time (min). Mean Δ 0–120, mean change from baseline over the 2 hours (area under the change curve divided by 120 min). LSC, least significant change between two scans ($1.96 \times \sqrt{2} \times \sigma$). Rise within LSC: positive mean change but no time point beyond the LSC. Last column: response to the first 400 mL electrolyte drink in the 12-hour study (mean of +30 and +60 min minus pre-drink score).

## 3.5 Stability and time in range

Similar mean scores hid very different trajectories (Fig. 6). Participant 1 on the water day, Participant 3 on the electrolyte day and Participant 4 on the water day had mean scores within 2 points of each other (51.9, 52.7 and 53.6). Participants 1 and 4 also had almost the same spread (SD 4.9 and 5.3 points; range about 20 points), yet Participant 1 fell below 50 eight times and spent 60% of the day in range, whereas Participant 4 fell below 50 four times, spent 80% of the day in range and stayed at or above 50 for 11 consecutive scans. Participant 3's day, with the same mean, spanned 36 points (34.2 to 70.5), with dips to 34–38 points before drinks.

Across the ten days, mean scores ranged from 45.1 to 61.1, whereas time in range varied from 16% to 100% and stability from 4.5 to 9.1 points (Fig. 7A, B; Table 4), even though every participant followed the same 2.0 L fluid schedule. A single scan taken at a random time disagreed with the day's predominant state with a probability of 0–40% (mean 19%), that is, about one reading in five overall and up to two in five on some days. A single reading or a daily fluid-volume target therefore captured only part of how hydration actually behaved over the day.

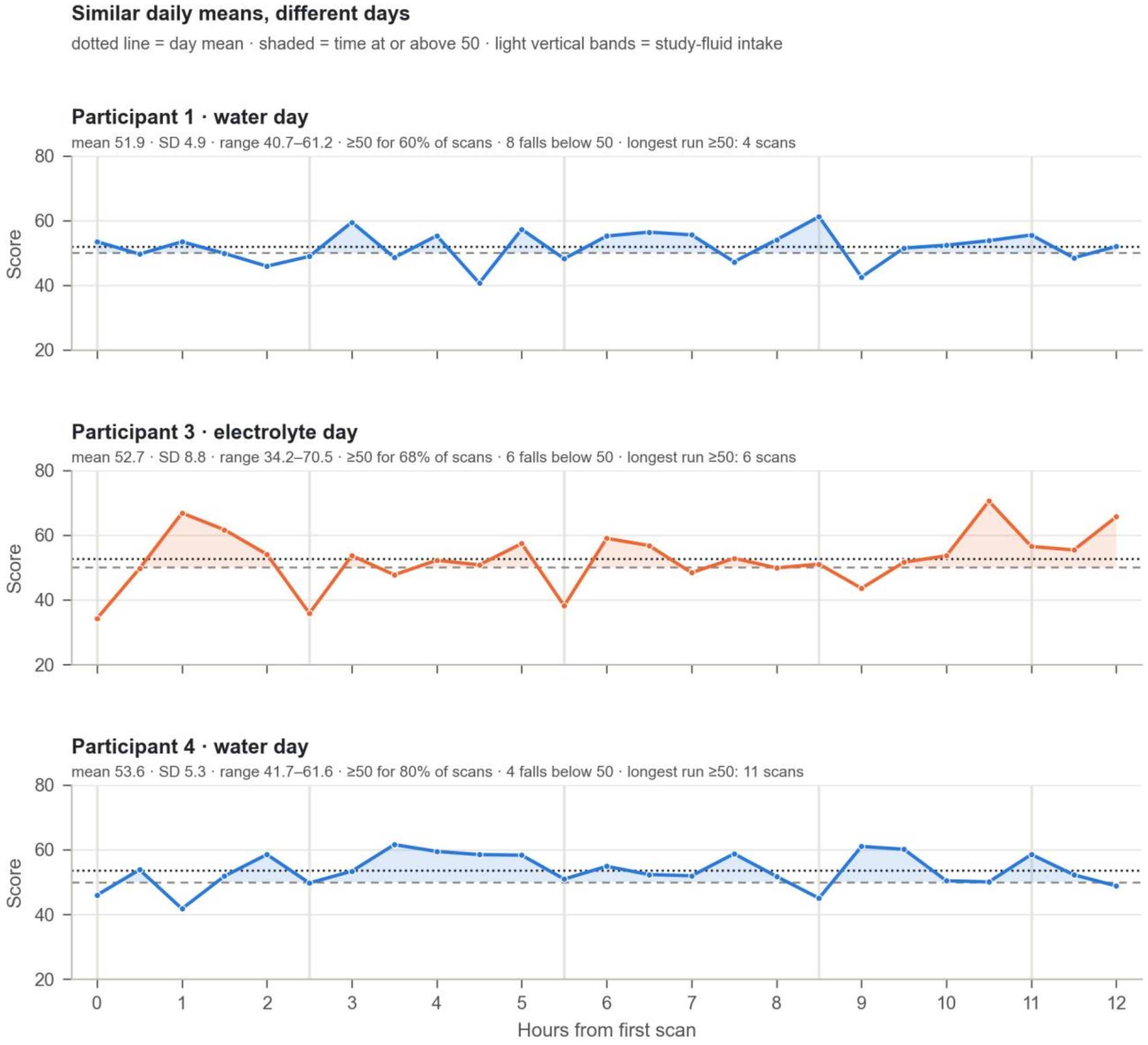

**Fig. 6.** Similar daily means, different days. Three days with mean scores within 2 points of each other but different spread, falls below 50 and time in range. Dotted line: day mean; shading: time at or above 50; light vertical bands: study-fluid intake.

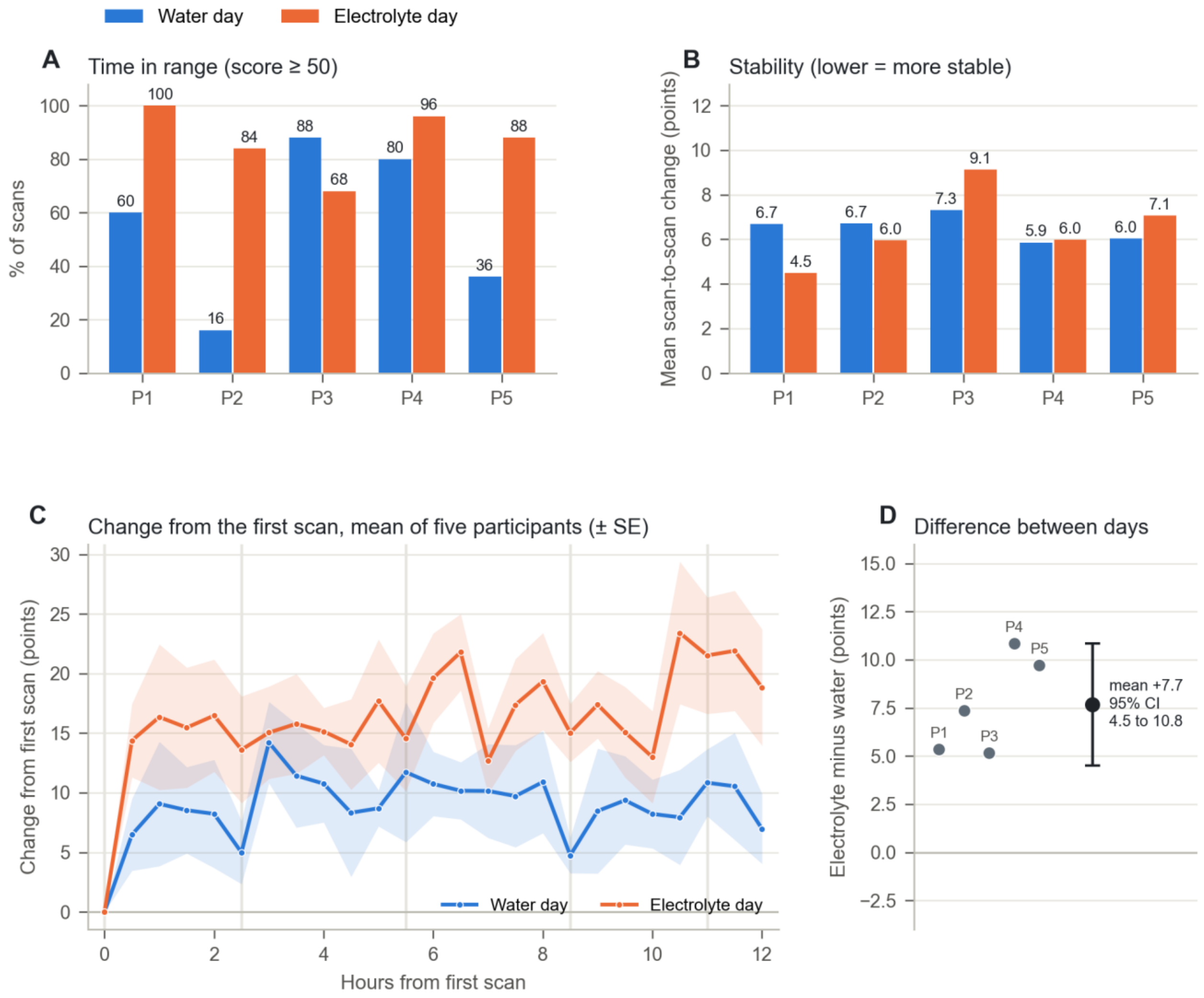

**Fig. 7.** Time in range (A; percentage of scans with a score of 50 or above) and stability (B; mean absolute change between consecutive scans) for each participant and day, under an identical 2.0 L fluid schedule. (C) Change from the first scan of the day, averaged across the five participants, on the water and electrolyte days (shading: ± 1 SE; vertical bands: study-fluid intake). (D) Each participant's difference between the two days (electrolyte minus water) in the mean change from the first scan, with the mean and 95% confidence interval.

## 4. Discussion

### 4.1 Main finding

The facial-video hydration biomarker can be used longitudinally, not only to detect a before-and-after change in state. Over 12-hour days, the score showed a low point in the morning after the overnight period without fluid, rose after the first drink whenever the day began below 50, and was higher after drinking than its own distribution would predict by chance. The response depended on the state in which fluid was taken: drinks taken below 50 were consistently followed by a rise, whereas drinks taken when the score was already at or above 50, which included most of the later 250–300 mL drinks, were not. This is what physiology predicts. Body water makes up roughly half to 60% of adult body mass [9], so a 250–300 mL drink changes total body water by less than 1%; when body water is already replete, a hypotonic fluid load suppresses arginine vasopressin and is largely excreted within hours, whereas after a deficit more of the ingested fluid is retained [4, 11, 27]. The pattern of consistent rises after drinks taken in deficit and little change after drinks taken in range therefore supports the physiological plausibility of the longitudinal signal (Objective 1).

The difference between the two study days is consistent with established fluid physiology, although the fixed order of the days requires caution. Beverages with a higher content of sodium and potassium, such as oral rehydration solution and milk, are retained better than plain water, with less urine produced over the following hours [11], and in euhydrated adults drinks containing 27 or 52 mmol/L of sodium were retained better than drinks containing 7 or 15 mmol/L [28]. In the present data, the electrolyte-day advantage appeared after the first drink and persisted between the later drinks, where better retention would be expected to show, whereas the immediate responses to later drinks were similar on both days. The drink used here (sodium 12.2 mmol/L) also provides potassium, calcium, magnesium and carbohydrate, so the size of any effect is likely to depend on drink composition, and a randomized-order comparison that includes drinks of different sodium content is needed to confirm it.

### 4.2 Hydration is personal

Different individuals showed different baseline patterns and different responses to similar fluid inputs (Objective 2). Under an identical schedule, time in range differed from 16% to 100% of the day; after the same 500 mL drink, the peak came at 30, 60 or 90 minutes and differed about three-fold in size; and the change needed to exceed measurement variation differed between participants, while their average scores were similar. The differences lay in dynamics rather than in level. Several physiological factors could contribute: the size of the overnight deficit, gastric emptying and absorption, whose speed varies between individuals [10, 29], renal water and sodium handling, body composition, menstrual-cycle effects on fluid regulation [23], and health conditions and medication such as those of Participant 3. Measurement factors could also contribute, including how well a particular face, skin type and device are captured. With five participants these sources cannot be separated. Either way, the result has the same practical consequence: the same score change cannot be interpreted identically for everyone.

The individual patterns were consistent with participants' characteristics, although with five participants they are illustrative only. Participant 1, the only participant with a normal body mass index and no health conditions, was the only one to begin both 12-hour days at or above 50, had the smallest scan-to-scan variation, and showed the largest early response to the standardized drink (+14.7 points at 30 minutes, beyond the personal least significant change). Participant 2, who had obesity, had the lowest morning scores and spent only 16% of the water day in range, in line with the population association between higher body mass index and inadequate hydration [30]. Participant 3, who had chronic kidney and lung disease and took regular medication, showed the most variable scores and was the only participant whose electrolyte day was not higher than the water day.

### 4.3 Toward personal calibration

Individual response curves may eventually allow hydration recommendations to be personalized (Objective 3). Three elements of personal calibration emerged from these data. The first is a personal measure of variation: the least significant change ranged from 14.2 to 21.4 points, so a 15-point rise is a clear signal for one person and within measurement variation for another. The second is a personal response curve to a standardized challenge, which in this study differed in magnitude, timing and persistence. The third is the state in which fluid is taken: the same drink was followed by a rise when the score was below 50 but not when it was already in range, so a personal protocol would time fluid to the person's measured state rather than to the clock alone. The 2-hour challenge and the comparable first drink of the 12-hour study did not agree closely, which indicates that response curves need to be measured under standardized conditions and interpreted against the starting state. Before response curves can inform dosing, their reproducibility must be established with repeated challenges in larger samples (Fig. 8).

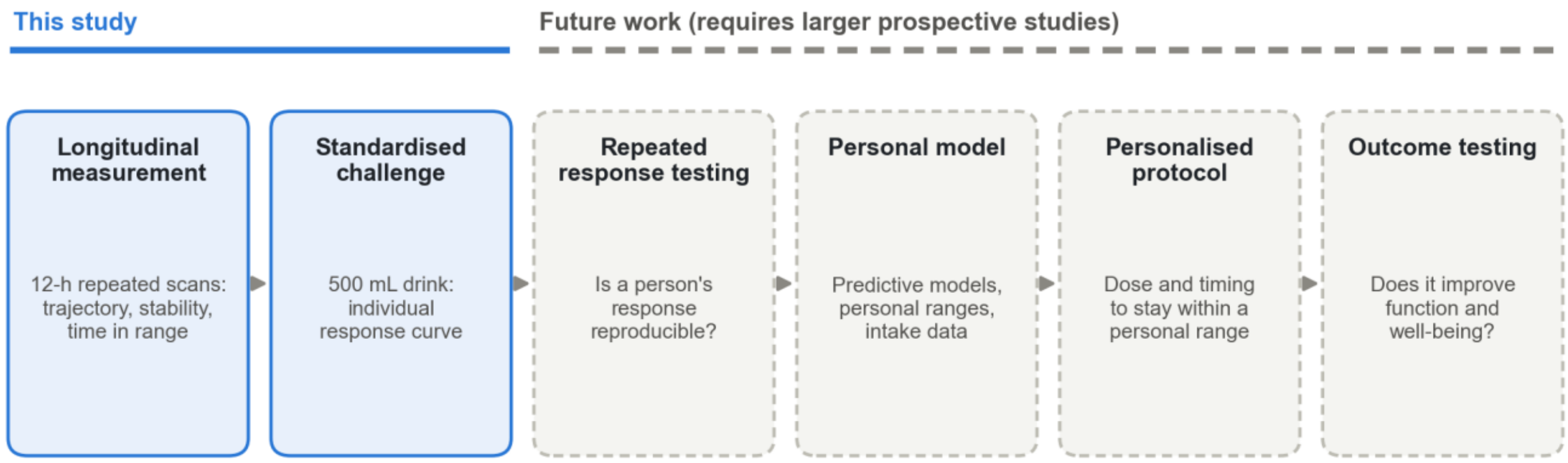

**Fig. 8.** Pathway from longitudinal measurement to personalized hydration protocols. Solid boxes: addressed in this study. Dashed boxes: future work requiring larger prospective studies.

## 4.4 Importance of trajectory and stability

The finding that days with similar mean scores can differ substantially in spread and time in range suggests that hydration should be considered as a dynamic range rather than a single measurement (secondary objective). Continuous glucose monitoring offers a precedent: time in range and glycaemic variability are reported alongside mean glucose because they carry information that the mean does not [24]. In the present data, a single scan disagreed with the day's predominant state in about one case in five (up to two in five on some days), and the same daily fluid volume produced time in range from 16% to 100%. The threshold of 50 used here is an operational boundary of the score, not an outcome-validated target, and appropriate personal ranges remain to be defined.

## 4.5 Technical and biological variability

Occasional implausible readings occurred. One scan (0.4% of 250), in Participant 1 on the water day (40.7 points at 4 h 30 min), departed from both neighbouring scans by more than the participant's least significant change while the neighbours agreed with each other; a change of body water of that size and reversal within one hour is not physiologically plausible, so it is more likely technical. All scans were retained in the analysis. A signal-quality gate and a repeat-scan rule would reduce the influence of such readings in practice.

Participant health and context also differed. Participant 3's chronic kidney condition, lung disease and antihypertensive medication alter fluid and electrolyte handling and may affect skin perfusion; Participant 3 was also the only participant whose electrolyte day was not higher than the water day. Participant 5's migraine episode occurred on the water day, on which every pre-drink score was below 50. These observations are descriptive and do not establish cause.

rPPG-based measurement has known limitations: it is sensitive to lighting, motion, camera and video compression, and its accuracy can vary with skin type [8, 25]. In addition, the morning measurements coincide with the cardiovascular changes that accompany waking, such as the morning rise in blood pressure [26], which could influence rPPG-derived signals independently of fluid intake. Finally, drink responses calculated from a single pre-drink scan are susceptible to regression toward the mean, particularly in participants with high scan-to-scan variation; the permutation test was chosen because it accounts for this, and the analysis by pre-drink score compared each group of intakes with the same selection in the shuffled data for the same reason.

### 4.6 Limitations

- **Small sample.** Five participants allow detailed individual description but not population inference.
- **Exploratory design.** Analyses were descriptive, p-values were not adjusted for multiple comparisons, and some analyses (the dependence of responses on the pre-drink score and the comparison of the two study days from their first scan) were defined after inspection of the data; they require confirmation in a prospective study.
- **Heterogeneous participants.** Body size and health status varied widely, which illustrates real-world diversity but limits comparison between individuals.
- **Need for repeat studies and larger validation cohorts.** Each participant completed one 2-hour challenge, without a no-drink control, so the reproducibility of individual response curves is unknown. The water day always preceded the electrolyte day and participants knew which drink they received, so order, expectation and drink effects cannot be separated. No concurrent biochemical reference (for example plasma osmolality or urine indices) was collected, sampling every 30 minutes cannot resolve faster changes, and health context was self-reported.

### 4.7 Future direction

The next steps follow directly from these limitations (Fig. 8):

- **Larger calibration studies** applying the standardized challenge to more participants, and a randomized-order comparison of water and electrolyte drinks of different sodium content.
- **Repeated response testing** to establish whether an individual's response curve is reproducible, including a no-drink control.
- **Predictive modelling** of individual trajectories from occasional scans and known fluid intake.
- **Personalized ranges** defined from each person's own baseline and variation.
- **Smart-bottle integration** to record fluid intake automatically alongside scans.
- **Outcome testing** in a future prospective study to determine whether maintaining a personal hydration range improves cognitive, functional and well-being outcomes.

## 5. Conclusion

The facial-video hydration biomarker can be used to characterize hydration longitudinally. Hydration trajectories differ between individuals, and standardized fluid challenges reveal different personal response patterns. These findings provide an early foundation for personalized hydration monitoring and future individualized hydration protocols.

**Author information.** All authors are affiliated with Hii.Health, Singapore.

**Competing interests.** Hii.Health develops the facial-video digital hydration biomarker evaluated in this study.